# Oracle Bone Script Similiar Character Screening Approach Based on Simsiam Contrastive Learning and Supervised Learning

Xinying Weng[1, a*] Weng
College of Computer and Information Engineering
Henan Normal University
Xinxiang, Henan, China
* Corresponding author: [a]wengxinying2021@163.com

Yifan Li[2, b] Li
College of Computer and Information Engineering
Henan Normal University
Xinxiang, Henan, China
[b]e-mail: 2108524052@stu.htu.edu.cn

Shuaidong Hao[3, c] Hao
College of Computer and Information Engineering
Henan Normal University
Xinxiang, Henan, China
[c]e-mail: 443461960@qq.com

Jialiang Hou[4, d] Hou
College of Computer and Information Engineering
Henan Normal University
Xinxiang, Henan, China
[d]e-mail: 2132226391@qq.com

*Abstract*—This project proposes a new method that uses fuzzy comprehensive evaluation method to integrate ResNet-50 self-supervised and RepVGG supervised learning. The source image dataset HWOBC oracle is taken as input, the target image is selected, and finally the most similar image is output in turn without any manual intervention. The same feature encoding method is not used for images of different modalities. Before the model training, the image data is preprocessed, and the image is enhanced by random rotation processing, self-square graph equalization theory algorithm, and gamma transform, which effectively enhances the key feature learning. Finally, the fuzzy comprehensive evaluation method is used to combine the results of supervised training and unsupervised training, which can better solve the "most similar" problem that is difficult to quantify. At present, there are many unknown oracle-bone inscriptions waiting for us to crack. Contacting with the glyphs can provide new ideas for cracking.

*Keywords- oracle decipher; ResNet-50; simsiam; RepVGG, homonym select*

## I. Introduction

The study of similar forms of characters in ancient scripts, such as oracle bone inscriptions, faces major challenges due to their diverse styles and uncertain forms [1]. This paper explores a new approach that integrates supervised and unsupervised learning techniques to improve recognition of similar morphological characters in Oracle [2]. By combining these methods, the proposed model aims to enhance the accuracy and efficiency of character recognition tasks. Powerful feature extraction and classification using deep learning models such as the RepVGG architecture are essential for distinguishing subtle changes in ancient texts [3]. In addition, on the basis of traditional methods, this study incorporates advanced image processing techniques and utilizes large-scale data sets [4]. In summary, this paper presents a comprehensive framework that contributes to the field of digital humanities [5]. Besides, a causal contrastivemechanism is developed to decouple the causality-driven and non-causal components of the discriminative features, suppress the non-causal ones, and hence improve the model stability and generalization [6]. SimSiam's simple Siamese network architecture for unsupervised visual representation learning [7].

## II. System Design

After inputting the original images, we first apply image enhancement techniques based on grayscale image processing during the data preprocessing stage. This includes employing gamma transformation and histogram equalization theory algorithms to effectively enhance the source images, making the information features more prominent in the processed images. Subsequently, through technical projection, we obtain secondary processed images. These images then undergo feature extraction of oracle bone script character data using a deep convolutional neural network reparameterized with the RepVGG model structure. Finally, the feature vector of the images is inputted into a dataset feature vector set under the SimSiam framework for similarity comparison and matching. This process achieves the selection, recognition, and classification of oracle bone script look-alike characters. This paper aims to develop a comprehensive model for selecting oracle bone script look-alike characters using the open-source HWOBC dataset.

### A. Data Collection Details

This paper aims to develop a comprehensive model for selecting oracle bone script look-alike characters using the open-source HWOBC dataset. Named "SimSiam-based and supervised learning fused oracle bone script look-alike character selection," the approach involves enhancing sample images post dataset acquisition to extract more prominent visual features. Supervised model and unsupervised model methods are used to train separately. Finally, the fuzzy comprehensive evaluation method is used to evaluate the similarity of the training effect of each model, which can enhance the credibility of the results in many ways , improve the work efficiency of oracle researchers

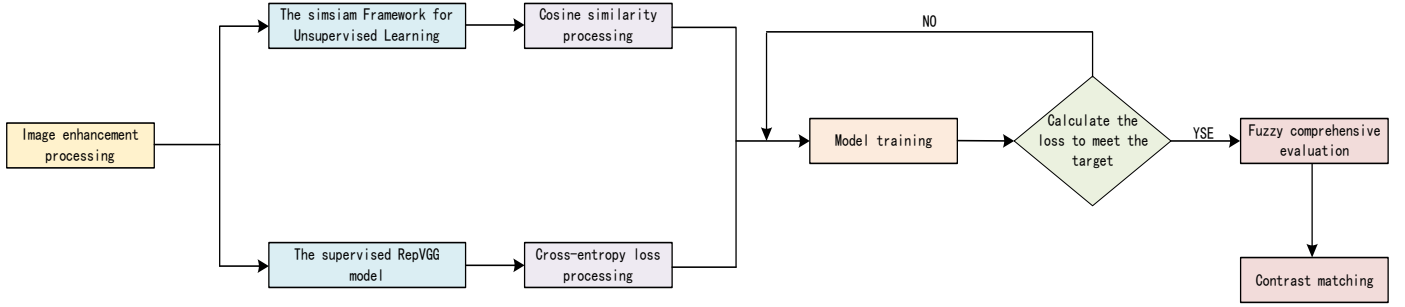

Figure 1. Structure diagram of oracle shape near-character recognition

and improve the accuracy of character recognition.

*B. Model Selection*

For our task of filtering Oracle Bone Script similiar characters, we have a large amount of labeled data, and the combination of supervised loss function and self-supervised learning objective function can improve the fit of the model. In this case, we use the weighted method to synthesize the training results of the two models to obtain the maximum result, and evaluate the size of the similarity in turn, so as to achieve better training effect. The overall process design is shown in Figure 1.

### III. IMAGE ENHANCEMENT TECHNIQUES BASED ON GRAYSCALE IMAGE PROCESSING

First, acquire the oracle bone script image source, as shown in Figure 2:

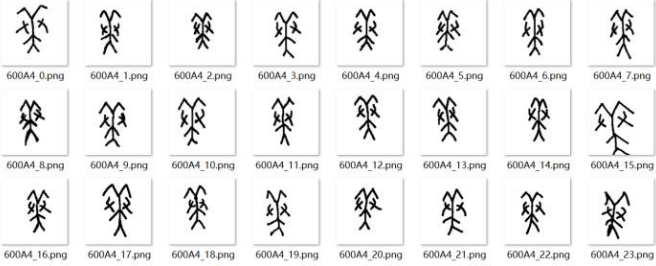

Figure 2. Part of the oracle bone image source

Then the image enhancement process is carried out to improve the single feature information in the oracle-bone inscriptions image, making the features more obvious, the feature extraction is more recognizable, and the results are easier to distinguish.

The word "Si" in Oracle Bone Script is selected as an example, and Figure 3 is the original image, and then the data enhancement process is carried out through the following steps.

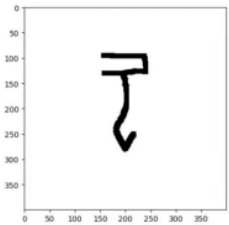

Figure 3. The original image of the character "Si" in oracle bone inscriptions

*A. Multi-angle rotation*

Rotation processing is an effective method for image data augmentation. In the data preprocessing stage, random rotation operation is introduced to increase the diversity and generalization ability of the training samples by rotating the image, so as to further increase the diversity of the training data.

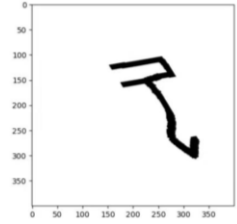

Figure 4. Rotate the processed image randomly

During training, the rotation Angle of each image can be chosen randomly. This method helps the model learn features that are invariant to rotation and improves the robustness of the model. Figure 4 shows the result of data augmentation after the image is rotated.

*B. Histogram Equalization Algorithm*

- In a histogram, if grayscale levels are concentrated in the high-intensity region, low-intensity areas of the image are less distinguishable. Conversely, if grayscale levels are concentrated in the low-intensity region, high-intensity areas become less distinguishable. To make both high and low intensities easily distinguishable, the best approach is to transform the image so that the probability distribution of grayscale levels is uniform.

$$s = T(r), 0 \leq r \leq L - 1 \quad (1)$$

- The relationship between the grayscale level distribution before and after transformation is:

$$P_s(s) = P_r(r) \left| \frac{dr}{ds} \right| \quad (2)$$

- Expressed discretely:

$$s_k = (L - 1) \sum_{j=0}^{k} \quad (3)$$

$$p_r(r_j) = \frac{L-1}{MN} \sum_{j=0}^{k} \quad j, k = 0,1,2, \ldots, L-1 \quad (4)$$

## C. Gamma Transformation

Gamma transformation is used for image enhancement, improving details in darker areas. Simply put, it involves a non-linear transformation that makes the image's response to exposure intensity more similar to human visual perception. It corrects images that are either overexposed (too bright) or underexposed (too dark). If gamma > 1, the grayscale values in brighter areas are stretched, making them brighter, while darker areas are compressed, making them darker, resulting in an overall darkening of the image. Conversely, if gamma < 1, the grayscale values in brighter areas are compressed, making them darker, while darker areas are stretched, making them brighter, resulting in an overall brightening of the image.

$$V_{out} = AV_{in}^Y \qquad (5)$$

After some Gaussian transformation processing, we can obtain the desired enhanced image. This is shown in Figure 5:

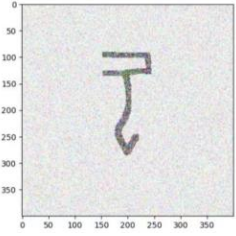

Figure 5. Text map after data augmentation

## IV. UNSUPERVISED NETWORK LEARNING BASED ON SIMSIAM

Based on the open-source HWOBC dataset of oracle bone inscriptions curated by Yinqi Wenyuan, we selected a subset of representative images for experimentation. Utilizing image augmentation techniques, we processed the images to extract additional information. We applied stop-gradient operations to restrict gradient updates on certain parameters, aiming to better focus on the distinctive features of oracle bone inscription images. This approach can potentially enhance the accuracy and stability of the model.

After training the ResNet-50 network, it can store feature vectors of the existing dataset [8]. Each time a query for similar oracle bone script images is made, the feature vector of the queried image is extracted and matched against the feature vectors in the existing dataset using the cosine similarity function to find the most similar images. In this matching process, a symmetric loss function is defined to measure the matching loss of all images and average the total loss over all images, with a minimum possible value of 1. An important component of the preliminary foundational work is the use of stop-gradient operations to prevent collapse. Therefore, the following settings are used for unsupervised pre-training.

The optimizer uses SGD for pre-training to update model parameters and minimize the loss function. There's no need for large-batch optimizers like LARS during pre-training. The learning rate is set to lr×BatchSize/256, where the base learning rate lr=0.05. A cosine decay schedule is employed for the learning rate, with weight decay set to 0.0001 and SGD momentum at 0.9. The introduction of averaged previous gradients smooths the direction of parameter updates during training. The residual module structure is shown in Figure 6.

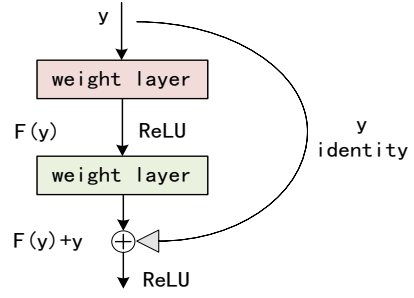

Figure 6. Residual module structure diagram

The projection MLP consists of 3 layers, each fully connected layer having 2048 dimensions. Each fully connected layer includes Batch Normalization (BN), including the output fully connected layer. The output fully connected layer does not include ReLU activation [9]. Using ResNet-50 as the default backbone, unsupervised pre-training was conducted on the specified oracle bone training dataset without supervised learning labels. The pre-trained feature encodings were saved to file and subsequently tested on the validation set, with the test results shown in Figure 7:

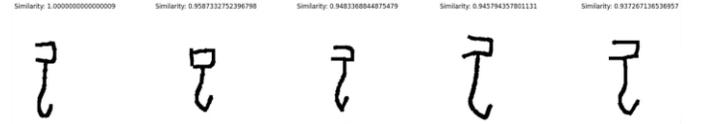

Figure 7. Unsupervised model results in oracle bone script image recognition

## V. LEARNING SUPERVISED MODELS BASED ON REPVGG

The convolution operation, batch normalization operation, ReLU activation function, adaptive average pooling operation and fully connected operation are used in the RepVGG network. Among them, the residual connection commonly used in ResNet is used in the Block module of RepVGG, and the feature extraction is completed by combining convolution, batch normalization and ReLU activation function. In the definition of the network, the structure idea of ResNet and DenseNet is borrowed, the network is divided into multiple stages, and the depth of the network is increased by stacking multiple identical modules, and the number of output channels is used to control the model size and computational complexity [10]. In the forward propagation process of the Block module of RepVGG, the element-wise addition operation is used to calculate the activation value of its ReLU. The formula can be expressed as follows.

$$out = ReLU(Conv(x,w) + BatchNorm(Conv(x,w)) \qquad (6)$$

RepVGG inherits the 3×3 convolutional kernel style of VGG, resulting in fast computation speed. It achieves this by integrating Conv2d and BN operations, and transforming branches with only BN into a single Conv2d operation, thereby consolidating the 3×3 convolutional layers on each branch into a single convolutional layer. The fusion process of convolutional layer and BN layer in RepVGG is shown in Figure 8.

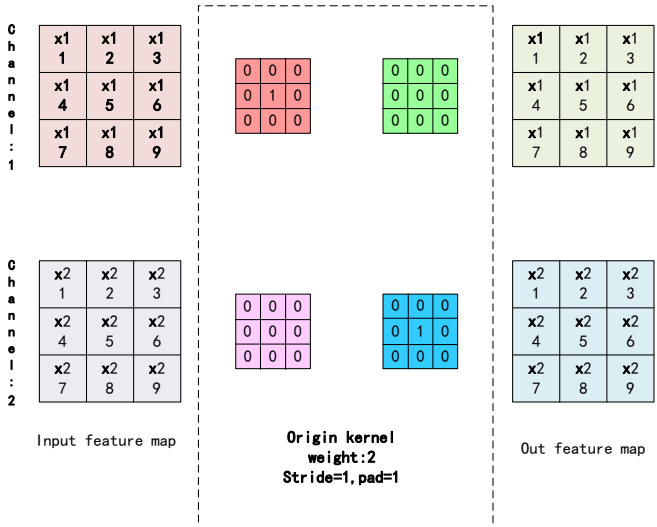

Figure 8. The convolutional layer is fused with the BN layer

In its design, RepVGG independently optimizes training and inference network structures: during training, it uses a high-precision multi-branch network to learn weights, while during inference, it employs a low-latency single-branch network. Through structural re-parameterization, weights from the multi-branch network are transferred to the single-branch network, ultimately reducing parameters and speeding up inference. The learning performance of RepVGG under supervised models is illustrated in Figure 9 [11]:

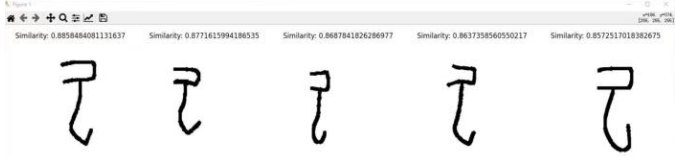

Figure 9. Supervised model results in oracle bone script image recognition

## VI. SIMILARITY ANALYSIS

Fuzzy comprehensive evaluation ensures comprehensive analysis of structured training outcomes and intrinsic data patterns, thereby enhancing the reliability and depth of similarity assessments. Similarity analysis integrates results from both unsupervised and supervised models, with equal weights assigned to each (0.5 each) [12]. This approach utilizes model outputs to comprehensively evaluate and compare similarities between data points or patterns, combining quantitative outputs with qualitative assessments for a deeper understanding of similarity.

## VII. CONCLUSIONS

In this study, a fusion of advanced techniques such as ResNet-50, RepVGG, and fuzzy comprehensive evaluation was proposed and put into the oracle-shaped near-character screening. It is well known that there may be some relationship between glyphics in human created scripts, and the study of glyphics is of great significance for understanding the structure and evolution of language and writing systems. At present, there are about 4500 unique characters in the unearthed oracle bone inscriptions, and about 2900 unrecognizable characters, accounting for about two-thirds of the total, which leaves huge research difficulties for the ancient writing workers. This paper innovatively combines Jiaguwen with digitization, through the combination of artificial intelligence technology, the use of supervised training and unsupervised training combined with fuzzy comprehensive evaluation method of the innovative method of fusion, so as to be able to screen out the shape near-character [13]. This study can provide important clues and ideas for the deciphering of oracle bone inscriptions and avoid a lot of repetitive workload for archaeological workers [14].